\documentclass[letterpaper, 10 pt, conference]{ieeeconf}  

\IEEEoverridecommandlockouts                              

\usepackage{graphics} 
\usepackage{epsfig} 
\usepackage{amsmath} 
\usepackage{graphicx}
\usepackage{adjustbox}
\usepackage{tabularray}
\usepackage{multirow}
\usepackage{array}
\usepackage{stfloats}

\title{\LARGE \bf
Object-Pose Estimation With Neural Population Codes
}

\author{Heiko Hoffmann$^{1}$ and Richard Hoffmann$^{2}$
\thanks{*No funding was received for this work.}
\thanks{$^{1}$Magimine, LLC,
        Simi Valley, CA 93065, USA.
        {\tt\small heiko@magimine.com}}%
\thanks{$^{2}$Magimine, LLC.
        {\tt\small rhoffman@caltech.edu}}%
}

\begin{document}

\maketitle
\pagestyle{plain} 

\begin{abstract}
Robotic assembly tasks require object-pose estimation, particularly for tasks that avoid costly mechanical constraints. Object symmetry complicates the direct mapping of sensory input to object rotation, as the rotation becomes ambiguous and lacks a unique training target. Some proposed solutions involve evaluating multiple pose hypotheses against the input or predicting a probability distribution, but these approaches suffer from significant computational overhead. Here, we show that representing object rotation with a neural population code overcomes these limitations, enabling a direct mapping to rotation and end-to-end learning. As a result, population codes facilitate fast and accurate pose estimation. On the T-LESS dataset, we achieve inference in 3.2 milliseconds on an Apple M1 CPU and a Maximum Symmetry-Aware Surface Distance accuracy of 84.7\% using only gray-scale image input, compared to 69.7\% accuracy when directly mapping to pose.
\end{abstract}

\section{INTRODUCTION}

Robotic manipulation requires object pose estimation; particularly, for assembly, we need accurate object orientation for part alignment. Traditionally, part alignment has been solved using mechanical constraints, where the orientation of a part is physically restricted at the pick-up location or where the gripper is engineered in a way to force a certain part orientation. These mechanical constraints, however, increase setup time and costs. Moreover, they cannot prevent failure, e.g., due to part jamming. 

Therefore, optical sensing of object pose is preferred. Pose consists of both object orientation and location. Here, orientation presents the primary challenge, as in many manufacturing settings, the mathematical representation of object rotation is not unique. For symmetric parts, multiple rotation matrices or quaternions can represent the same pose. This ambiguity creates a problem for learning algorithms, such as those mapping images to poses, since a neural feedforward network requires a unique target for each input and fails when encountering multiple possible outputs \cite{Hoffmann2005}.
 
As a solution, some works proposed to learn pose candidates, which are then down-selected on the sensory data \cite{wen2024,liu2024}. Alternatively, data can be mapped onto a feature space, e.g., with an autoencoder  \cite{sundermeyer2020}, which is then compared against the feature-space representation of a dense set of possible poses. But these indirect mappings have also disadvantages: they are slower, the algorithms are more complex, and they are less flexible in terms of mapping auxiliary information onto pose. 

Here, we propose a representation for object orientation that allows a direct mapping onto pose and end-to-end learning. We take inspiration from information-encoding in the mammalian cortex. There many variables, like object orientation \cite{Hubel1959}, have been found to be encoded by a population of neurons, where each neuron maximally responds to a preferred value, and the activation of the group of neurons resembles a probability distribution of the encoded variable \cite{Pouget2000}. The population code can represent object symmetry and the corresponding pose ambiguity as multiple activation peaks in the code.

For each input image, we predict the entire population code of the object rotation. We tested our population code on the T-LESS dataset and compared against alternative pose representations using the same network architecture and training data. As a result, we observe significantly improved rotation accuracies with fast inference (3.2 msec on an Apple M1 CPU).

\section{RELATED WORK}

Related to our work are methods that predict a probability distribution of pose or predict multiple possible pose outcomes.

\subsection{Probability Over Rotations}

Pose ambiguity can be dealt with by estimating a probability distribution over the group of rotations in 3D, ${\bf R} \in {\bf SO(3)}$ \cite{gilitschenski2019,murphy2021}. The distribution may be represented parametrically, e.g., with a Gaussian mixture model \cite{gilitschenski2019} or non-parametrically \cite{murphy2021}. The former has the disadvantage of being hard to train \cite{gilitschenski2019}, e.g., due to local minima \cite{Hoffmann2005}.

In the non-parametric method by Murphy et al.,  the distribution is predicted implicitly, i.e., a neural network computes the probability $p({\bf R}, I)$ for any input matrix ${\bf R}$ and input image $I$ \cite{murphy2021}. The disadvantage of this strategy is that $p$ has to be normalized across a large sample of rotation matrices that all have to be fed through the network, resulting in a computationally expensive and slow training.

Given a normalized probability and a ground-truth rotation matrix ${\bf R}_o$, the loss function is defined as follows  \cite{murphy2021}
\begin{equation}\label{eq:prob_loss}
\mathcal{L} = - \log(p({\bf R}_o | I))\,.
\end{equation}

For inference, the rotation with maximum probability is chosen from a set of centers of an equivolumetric partitioning of {\bf SO(3)} \cite{murphy2021},

\begin{equation}\label{eq:prob_inf}
{\bf \hat R} = \underset{{\bf R} \in {\bf SO(3)}}{\arg \max}\, p({\bf R} | I)\,.
\end{equation}

\subsection{Multiple Pose Hypotheses}

Manhardt et al. train a network to predict multiple pose hypotheses to learn alternative (ambiguous) poses \cite{manhardt2019}. The network output is a fixed number $M$ of possible rotations. For training, the loss function combines a minimum loss $\mathcal{L}_{\mbox{\small min}}$, the minimum mean squared error (MSE) between the ground truth and the $M$ outcomes, with an average loss $\mathcal{L}_{\mbox{\small avg}}$, which is the average MSE across the $M$ outcomes,

\begin{equation}\label{eq:M-loss}
\mathcal{L} = (1-\epsilon \frac{M}{M-1}) \mathcal{L}_{\mbox{\small min}} + \epsilon \frac{M}{M-1}  \mathcal{L}_{\mbox{\small avg}}\,,
\end{equation}
where the parameter $\epsilon$ is gradually reduced from 0.05 to 0.01.

Inference is more complicated since we do not know a probability for each of the $M$ outcomes. Instead, the outcomes are clustered, and the median for each cluster is computed. The results can be compared against the input image to find the best fit  \cite{manhardt2019}. Though this strategy involves again an extra selection step as mentioned above.

\section{METHODOLOGY}

The key element of our new method is to represent object orientation with a neural population code. This code can be the last layer in any neural network architecture. In the following, we describe the population code and the specific architecture that we used in our real-world experiments. 

\subsection{Orientation Code}

To encode the orientation with a neural population code, we first transform the rotation matrix into a rotation axis and a rotation angle. For the axis, the preferred values of each neuron are arranged on a sphere. To obtain a near uniform distribution of axes, we compute a Fibonacci lattice on a sphere \cite{dixon91}. For the rotation angle, the preferred values are arranged in a circle. To combine axis with angle, our population code has for each point on the Fibonacci sphere a circle of neurons for the rotation angle. So, the space to represent the orientation is a direct product of a sphere and a circle. 

Given a rotation axis and angle, we activate all neurons $\{ i \}$ using Gaussian tuning curves,

\begin{equation}
a_i = \exp\left(-\frac{d\theta_i^2 + d\phi_i^2}{2 \sigma^2}\right)\, ,
\end{equation}
where $d\theta_i$ is the angle between the encoded axis and the preferred axis on the sphere for neuron $i$, $d\phi_i$ the angle between the encoded angle and the preferred angle of neuron $i$, and $\sigma$ is the tuning width, here $20^\circ$.
 
\begin{figure}[h]
\centering
{\small Input Image\hspace{2.3cm}Population Code}\\ \vspace{2mm}
\includegraphics[width=0.48\textwidth]{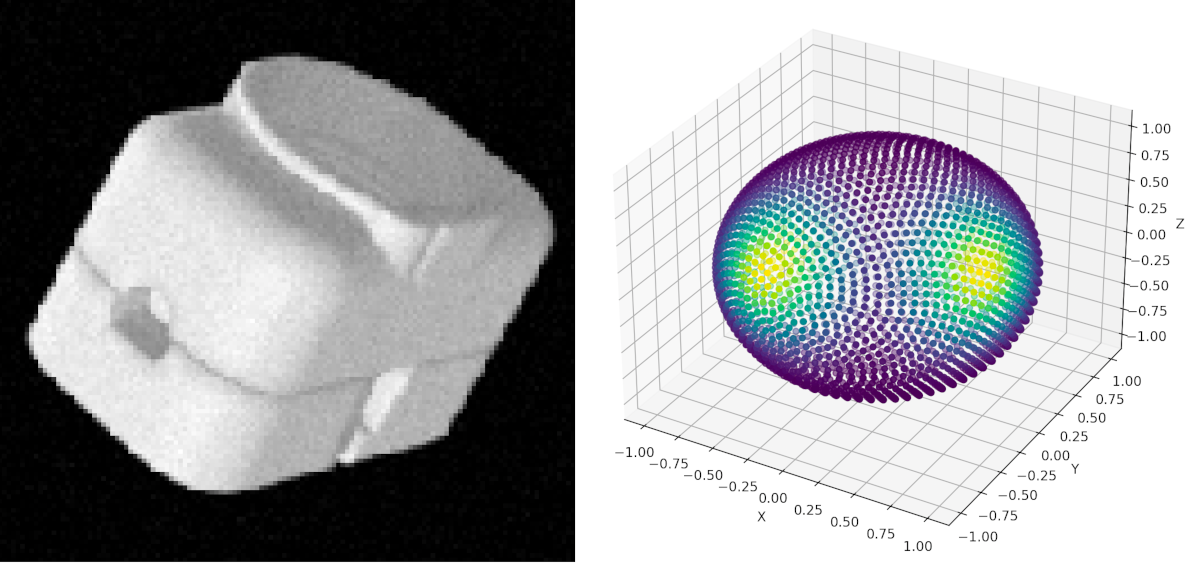}\\ \vspace{1mm}
\includegraphics[width=0.48\textwidth]{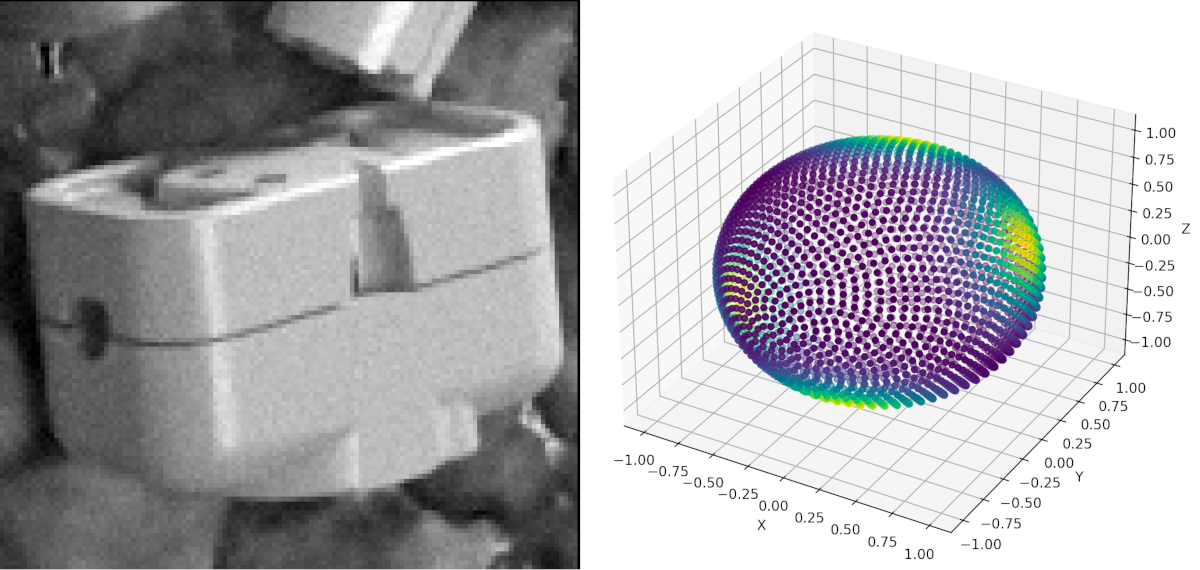}
\caption{The population-code for symmetric objects has multiple peaks. The neural activation is color coded (yellow: high; blue: low). Here, showing only the Fibonacci sphere for the rotation axis.}\label{fig:code}
\end{figure}

 To compute the target population code from a ground truth transformation matrix, ${\bf R}_o$, we compute activations for all symmetry transformations of  ${\bf R}_o$,  
 \begin{equation}
 {\bf R'}_k = {\bf R}_o {\bf R}_{\mbox{\small sym}}^k\,,
 \end{equation}
where $\{ {\bf R}_{\mbox{\small sym}}^k\}$ is the set of symmetry transformations for a given object (in a manufacturing setting, the symmetry of an object is known a priori). The target population code is the sum of all activations $a_i$ over all $k$ symmetry transformations (Fig. \ref{fig:code}).
 
 The transformation from rotation matrix to axis/angle, ${\bf r}$/$\phi$, is not unique since $-{\bf r}$ and $2 \pi - \phi$ is an equivalent axis/angle combination. Therefore, for each symmetry transformation, we compute the tuning-curve activations also for $-{\bf r}$ and $2 \pi - \phi$ and add those to the target activations (resulting in four peaks in Fig. \ref{fig:code}).
 
For objects with a symmetry axis, which would result in infinitely many equivalent poses, we compute a variant of our population code: since the population code for the rotation angle would show uniform activations, we simplify the code and omit the rotation angle neurons and just use the neurons encoding the rotation axis on the Fibonacci sphere. Here, the code is computed simply as
\begin{equation}
a_i = \exp\left(-\frac{d\theta_i^2 }{2 \sigma^2}\right)\, ,
\end{equation}
 without having to superimpose activations for symmetries. So, our population code can leverage the rotation symmetry.
 
For inference, we predict the entire population code and choose the neuron with the highest activation,
\begin{equation}
j = \underset{i}{\arg \max}\, a_i\,.
\end{equation}
Neuron $j$ corresponds to a rotation axis and angle, which we convert into a rotation matrix. For objects with symmetry axis, we pick a random angle. This inference is similar to (\ref{eq:prob_inf}) from the probabilistic method \cite{murphy2021}  with the crucial difference that we only need to evaluate our network once, resulting in a faster pose estimate.

\subsection{Network Architecture}

\begin{figure*}[!t]
      \centering
      \includegraphics[width=0.76\textwidth]{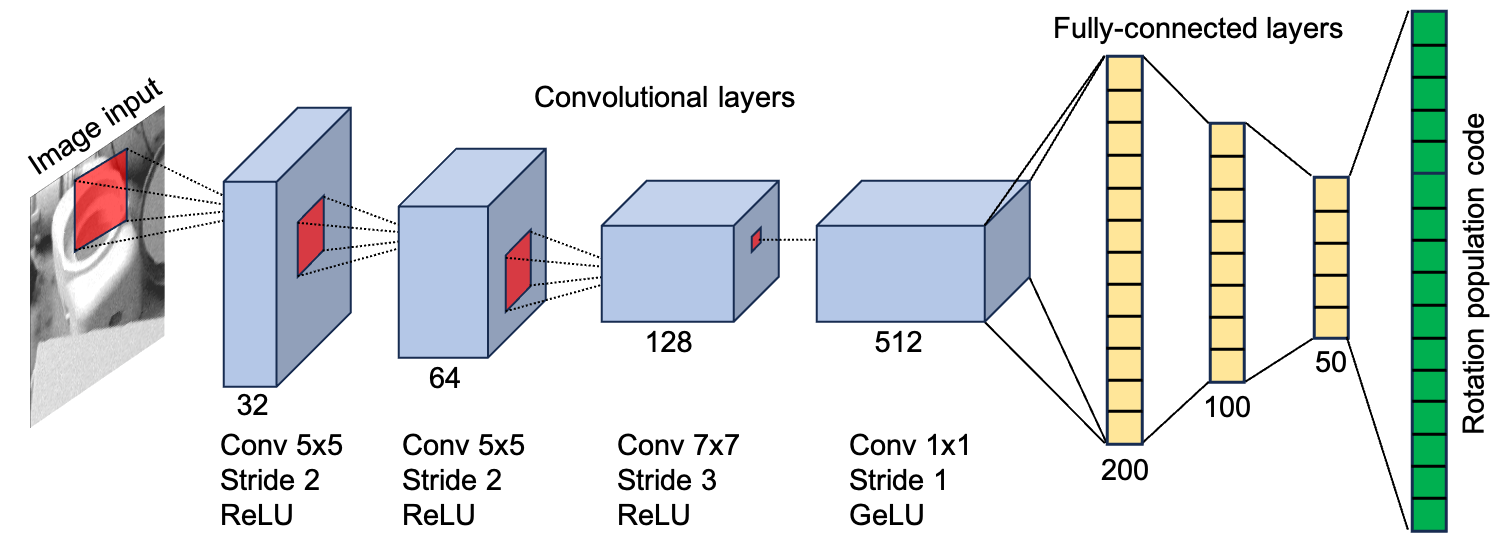}\vspace{-3mm}
      \caption{Our network architecture comprises a sequence of 4 convolutional blocks and 4 linear layers}\label{fig:arch}
 \end{figure*}

Our neural network maps gray-scale images of size 128x128 pixels onto a population code, which is a vector of size $n m$, where $n$ is the number of preferred axes and $m$ the number of preferred angles. Here, we used $n = 2,562$ and $m = 36$, matching the number of rotation candidates used in \cite{sundermeyer2020}.

The images are fed through four convolutional layers (see Fig. \ref{fig:arch} regarding the kernel size and number of features). The first three convolutional layers are followed by batch normalization and ReLU activation functions. This architectural choice was inspired by the Augmented Autoencoder \cite{sundermeyer2020}. The fourth convolutional layer with kernel size 1x1 is followed by a GeLU non-linear function. This architectural element was inspired by \cite{liu2022}, which demonstrated its benefit.

After the convolutional layers, the features are mapped through three hidden linear layers onto the population code, which itself acts like a linear layer. Each hidden layer is followed by a Leaky ReLU activation function.

\section{EVALUATION}

We evaluated our method on synthetic and real-world data.

\subsection{Synthetic-Data Experiment}

\begin{figure}[h]
    \centering
    {\small Bars}\\
    \includegraphics[width=0.7\linewidth]{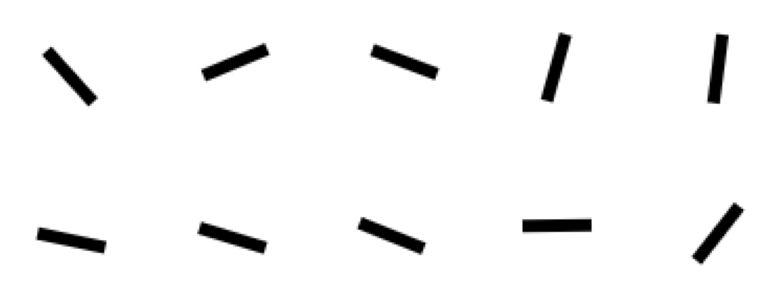}\\
    {\small Arrows}\\
    \includegraphics[width=0.7\linewidth]{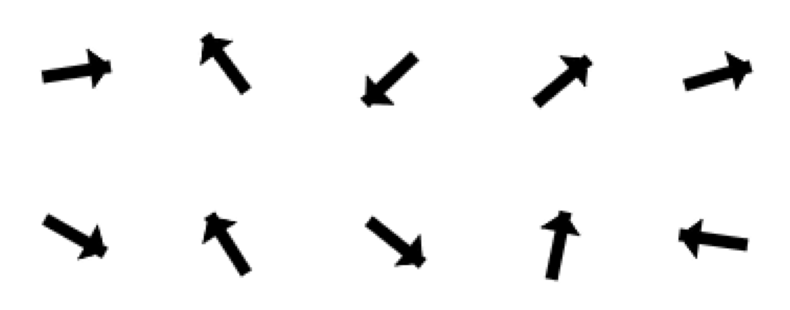}
    \vspace{-2mm}
    \caption{Random selection of training images for our synthetic datasets}
    \label{fig:bars}
\end{figure}

The synthetic-data experiment demonstrates the benefit of the population code. Here, we predict the one-dimensional orientation of shapes inside 64x64 pixels images (Fig. \ref{fig:bars}). As training data, we used either a set of randomly generated bar or arrow images (800/200 train/test split).

For the synthetic data, we used a simplified neural network comprising three convolutional layers (feature sizes: 32, 64, and 128) and 4 linear layers. All layers except for the output layer were followed by a ReLU activation function, and for the convolutional layers, we added max pooling (2x2).

For the output, we used either a population code vector of 36 neurons or a scalar orientation angle. The population code had preferred angles uniformly distributed in a circle in increments of $10^\circ$ and Gaussian tuning curves with width $20^\circ$. For the one-hot vector, we use the same output layer as for the population code.

We trained the network with MSE loss for single variable and population code. For the one-hot vector, we tested both MSE and Cross Entropy loss. In addition, tor training the population code, we compared two strategies: 1) having a single activation peak at the ground truth angle and 2) having two activation peaks $180^\circ$  apart for shapes with $180^\circ$ symmetry. 

We repeated train and test runs 10 times, including new train/test splits for each run. As a result, learning the population code with symmetry (two peaks) had the lowest error  (Tab. \ref{tab:bar_results}). Omitting symmetry during training resulted in a slightly larger error, but was not catastrophic. In contrast, learning failed when mapping directly onto the orientation angle for the symmetric bar.

\begin{table}[h]
    \centering
    \caption{Squared error of orientation angle for the arrows and bars datasets. For the Sym variant, the method was trained on the bars dataset with two possible angles as a target.}
    \vspace{-1mm}
    \label{tab:bar_results}
    \begin{tabular}{@{}lccc@{}}
        \hline
        \textbf{Method} &  \textbf{Arrows} & \textbf{Bars}\\
        \hline
        Single Variable & $237 \pm 69$ & $3597 \pm 423$\\
        One Hot (MSE) &  $112 \pm 229$ & $15.0 \pm 1.7$\\
        One Hot (Cross Entropy) & $13.8 \pm 1.2$ & $12.6 \pm 1.2$\\
        Population Code  & $\mathbf{8.6 \pm 0.5}$ & $17.8 \pm 2.3$\\
        Population Code (Sym)  & N/A & $\mathbf{8.7 \pm 0.5}$\\
        \hline
    \end{tabular}
\end{table}

\subsection{Evaluation Metrics}

For the real-world data, we computed six different metrics to evaluate the accuracy of estimating pose; specifically, we evaluated the rotation error, assuming a perfect translation estimate. We chose metrics that are suitable for symmetric objects. The first three metrics are used by the BOP: Benchmark for 6D Object Pose Estimation \cite{hodan2020}. The remaining three have been also used in the literature.

\vspace{2mm}
\noindent {\bf Maximum Symmetry-Aware Surface Distance (MSSD)}

Given a set of model vertices, $V_M$, the MSSD error is computed as

\begin{equation}
e_{\mbox{\small MSSD}} = \underset{{\bf R}_{\mbox{\tiny sym}} \in S_M}{\min} \underset{{\bf x }\in V_M}{\max} || {\bf \hat R}{\bf x} - {\bf R'}_o{\bf x} ||_2\, ,
\end{equation}
where $S_M$ is a set of symmetry transforms, s. th., ${\bf R'}_o = {\bf R}_o {\bf R}_{\mbox{\small sym}}$. The MSSD accuracy is the mean recall rate of $e_{\mbox{\small MSSD}} < \theta$ with $\theta$ ranging from 5\% to 50\% of the object diameter in steps of 5\% \cite{hodan2020}.

\vspace{3mm}
\noindent {\bf Maximum Symmetry-Aware Projection Distance (MSPD)}

The MSPD error is similar to the MSSD error, except that it takes into account only the projection onto the image plane,
\begin{equation}
e_{\mbox{\small MSPD}} = \underset{{\bf R}_{\mbox{\tiny sym}} \in S_M}{\min} \underset{{\bf x }\in V_M}{\max} ||{\bf P} ({\bf \hat R}{\bf x} - {\bf R'}_o{\bf x} )||_2\, ,
\end{equation}
where ${\bf P}$ is the projection onto the image plane. The MSPD accuracy is the average recall rate of $e_{\mbox{\small MSPD}} < \theta$ with $\theta$ ranging from $5w/640$ to $50w/640$ in steps of $5w/640$, where $w$ is the image width in pixels \cite{hodan2020}.

\vspace{3mm}
\noindent {\bf Visible Surface Discrepancy (VSD)}

The VSD error measures the mismatch in appearance between the model in the ground-truth pose and the same model in the estimated pose,
\begin{equation}
e_{\mbox{\small VSD}} =  \underset{x \in \hat V\cup V_o}{\mbox{avg}}\left\{ 
\begin{array}{ll}
0 & \text{if } x \in \hat V \cap V_o \land |\hat d(x) - d_o(x)| < \tau \\
1 & \text{otherwise}\,,
\end{array} 
\right. 
\end{equation}
where $\hat V$ and $V_o$ are the projected shapes of the model onto the image plane for estimate and ground truth, and $\hat d(x)$ and $d_o(x)$ are the estimated and ground-truth depths of the model at pixel $x$. The VSD accuracy is the average recall rate of $e_{\mbox{\small VSD}} < \theta$ with parameter $\tau$ ranging from 5\% to 50\% of the object diameter in steps of 5\% and $\theta$ ranging from 0.05 to 0.5 in steps of 0.05 \cite{hodan2020}.

\vspace{3mm}
\noindent ${\bf e}_{\mbox{\small\bf VSD}} {\bf < 0.3}$

In this variant, the accuracy is the ratio of errors below threshold 0.3 with $\tau = 20$. We adopt this metric since it was used by Sundermeyer et al for the augmented autoencoder \cite{sundermeyer2020}.

\vspace{3mm}
\noindent {\bf Visual Surface Similarty (VSS) }

VSS is a variant of VSD where $\tau = \infty$, i.e., the error in depth is ignored. This metric has been used for the multiple pose hypotheses method \cite{manhardt2019}.

\vspace{3mm}
\noindent {\bf Average Distance of Model Points - Indistinguishable Views (ADI) }

ADI is a variant of the Average Distance of Model Points metric, adapted for symmetric objects \cite{hodan2016}. The minimization takes into account ambiguous poses. 
\begin{equation}
e_{\mbox{\small ADI}} = \underset{{\bf x}\in V_M}{\mbox{avg}} \underset{{\bf y }\in V_M}{\min} || {\bf \hat R}{\bf y} - {\bf R}_o{\bf x} ||_2\, ,
\end{equation}
To compute the ADI accuracy, we computed the ratio of $e_{\mbox{\small ADI}}  < \theta$ with $\theta$ equal to 5\% of the object diameter. Here, we used a different threshold than in \cite{hodan2016}, who used 15\%, because with $\theta = 15$\% of diameter, all our experiments would show a 100\% accuracy.

\vspace{3mm}
\noindent {\bf Sensitivity to rotation error}

\begin{figure}[h]
      \centering
      \includegraphics[width=0.49\textwidth]{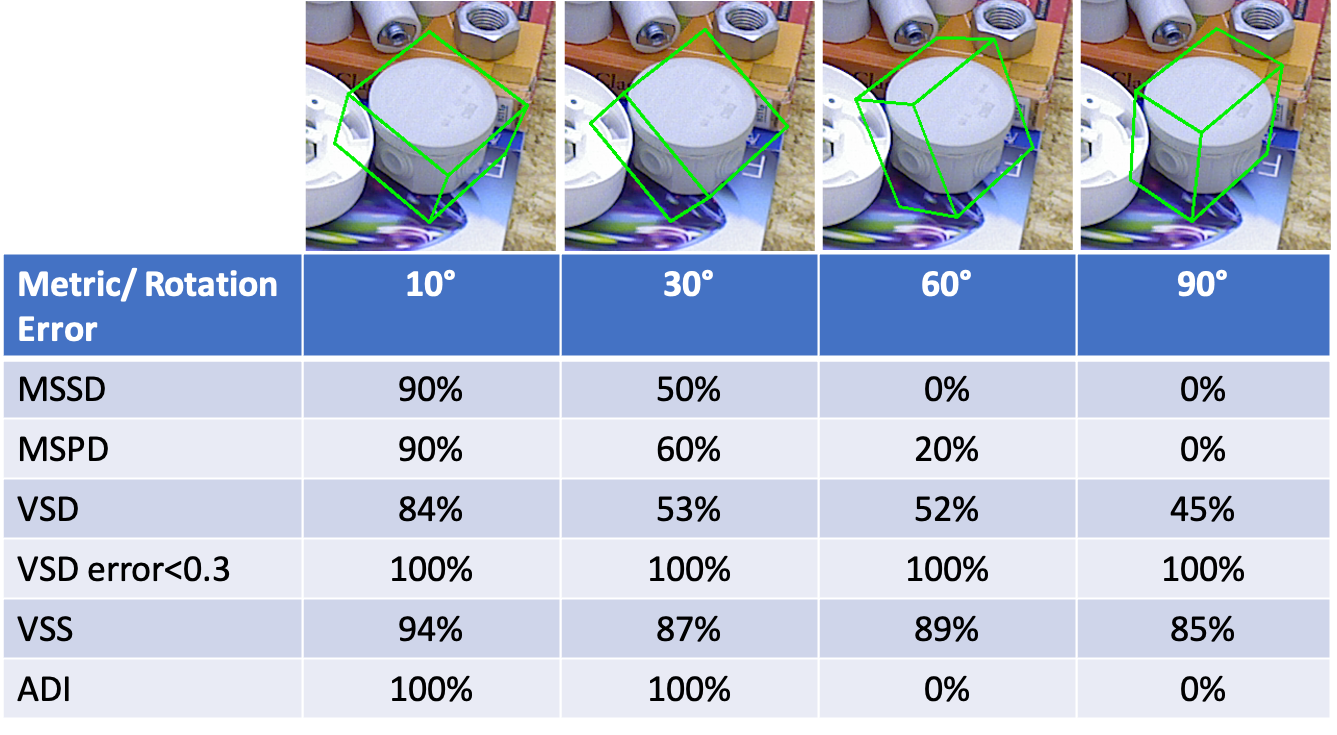}\vspace{-3mm}
      \caption{Example of sensitivity of metrics to errors in the rotation estimate}\label{fig:metrics}
 \end{figure}

We found large differences in sensitivity to rotation error between metrics (Fig. \ref{fig:metrics}). Particularly, for roughly spherical objects most metrics failed to reflect large errors. Moreover, only MSSD and ADI are invariant to the viewpoint since the other metrics compute a projection onto the image plane. To conclude, we found MSSD to be the most suitable metric to estimate rotation accuracy.  

\subsection{Comparison Methods}
 
On the real-world data, we compared our method with three other methods. All of them used the same network architecture, as described in Section III, and the same data augmentation. This ensures a direct comparison of the impact of the representations. In contrast, results across publications are difficult to compare because architecture and data augmentation significantly affect the outcomes but are often hard to replicate due to a lack of detailed information. This section describes the alternative methods.
  
\vspace{2mm}
\noindent {\bf Multiple Pose Hypotheses}

We adapted the multiple pose hypotheses method \cite{manhardt2019} to our network, predicting $M$ quaternions in the last layer (size: $4M$). For training, we used the loss in (\ref{eq:M-loss}), and for inference, we computed mean-shift clustering on the $M$ predicted quaternions \cite{manhardt2019} and picked the mean of the largest cluster.
 
 \vspace{2mm} 
\noindent {\bf Single Variable}

For objects with a symmetry axis, we can directly predict the three coordinates of the axis since there is no ambiguity. Here, the last layer of our network had only three elements, and for training, we used the MSE to the ground truth axis.

For objects with discrete symmetries, we computed the minimum loss to the closest target (one for each equivalent pose). Here, the last layer of our network was 6 dimensional for storing two columns of the rotation matrix {\bf R}. Learning two columns has been shown to be advantageous over other representations like quaternions due to the continuity of the  {\bf R}6 space \cite{zhou2019}. This representation was also used by one of the best performing methods for asymmetric objects \cite{wang2021}, and, as we will see, for asymmetric objects, mapping onto {\bf R}6 does indeed work well (Tab. \ref{tab:all_obj}). The entire rotation matrix can be reconstructed from the two columns using Gram-Schmidt orthonormalization. For training, we used the L1 loss.

 \vspace{2mm}
 \noindent {\bf One-Hot Vector}
 
Instead of treating the output layer as a population code, we used it as a one-hot vector, where each element/index represents a rotation. As common for one-hot vectors, we used the Cross-Entropy loss for training. For inference, we picked the index with the largest logit/probability.
 
\subsection{Real-World-Data Experiment}

We carried out our real-world experiments with the T-LESS dataset \cite{hodan2017}, which contains feature-less and mostly symmetric objects.  Table \ref{tab:all_obj} shows all 30 objects.

\begin{table*}[bp]
\caption{Population-code, one-hot vector, and single variable results (MSSD accuracy) for all 30 objects in the T-LESS dataset}\label{tab:all_obj}
\vspace{-1mm}
\centering
\resizebox{\textwidth}{!}{
\begin{tabular}{| l ||c|c|c|c|c|c|c|c|c|c|}
\hline
Object &\includegraphics[width=0.1\textwidth]{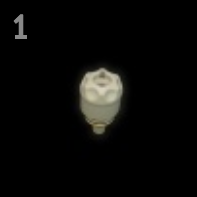} & \includegraphics[width=0.1\textwidth]{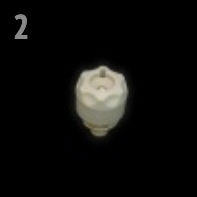} & \includegraphics[width=0.1\textwidth]{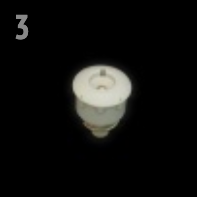} & \includegraphics[width=0.1\textwidth]{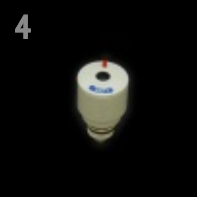} & \includegraphics[width=0.1\textwidth]{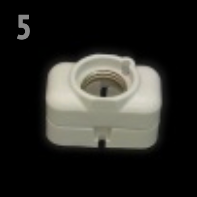} & \includegraphics[width=0.1\textwidth]{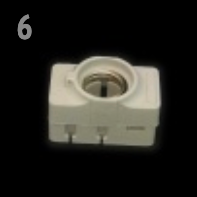} & \includegraphics[width=0.1\textwidth]{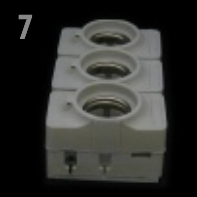} & \includegraphics[width=0.1\textwidth]{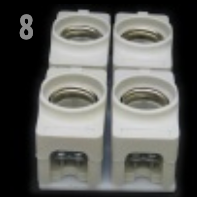} & \includegraphics[width=0.1\textwidth]{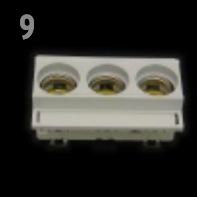} & \includegraphics[width=0.1\textwidth]{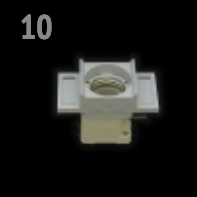}\\
\hline
Symmetries & Continuous & Continuous & Continuous & Continuous & 2 & 2 & 2 & 2 & 2 & 2\\
\hline
Trained on Objects & 1, 2 & 1, 2 & 3, 4 & 3, 4 & 5, 6 & 5, 6 & 7, 8 & 7, 8 & 9 & 10 \\
\hline
Test Instances & 818 & 465 & 397 & 619 & 194 & 100 & 250 & 150 & 249 & 145 \\
\hline
Single Variable & 80.4\% & 80.9\% & 90.3\% & 84.1\% & 66.0\% & 64.9\% & 21.9\% & 34.9\% & 58.7\% & 64.0\% \\
\hline
One Hot Vector & 71.0\% & 74.5\% & 82.1\% & 78.5\% & 69.0\% & 50.7\% & 41.5\% & 30.9\% & 41.6\% & 51.5\% \\
\hline
Population Code & \textbf{81.8\%} & \textbf{84.1\%} & \textbf{91.6\%} & \textbf{86.8\%} & \textbf{89.6\%} & \textbf{91.5\%} & \textbf{83.2\%} & \textbf{89.1\%} & \textbf{92.2\%} & \textbf{92.5\%} \\
\hline\hline

Object & \includegraphics[width=0.1\textwidth]{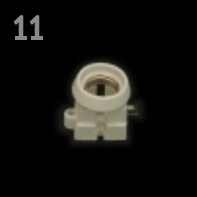} & \includegraphics[width=0.1\textwidth]{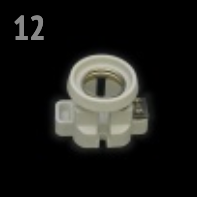} & \includegraphics[width=0.1\textwidth]{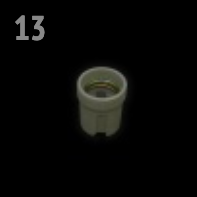} & \includegraphics[width=0.1\textwidth]{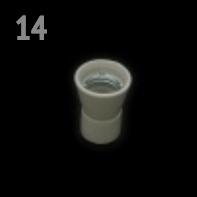} & \includegraphics[width=0.1\textwidth]{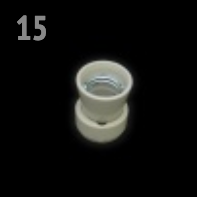} & \includegraphics[width=0.1\textwidth]{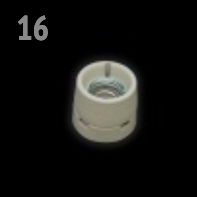} & \includegraphics[width=0.1\textwidth]{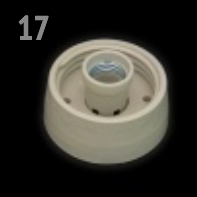} & \includegraphics[width=0.1\textwidth]{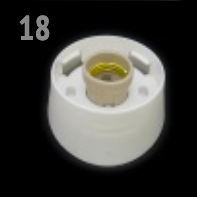} & \includegraphics[width=0.1\textwidth]{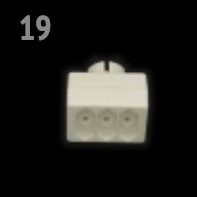} & \includegraphics[width=0.1\textwidth]{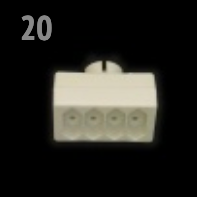} \\
\hline
Symmetries & 2 & 2 & Continuous & Continuous & Continuous & Continuous & Continuous  & None & 2 & 2\\
\hline
Trained on Objects & 11, 12 & 11, 12 & 13, 14 & 13, 14 & 15, 16 & 15, 16 & 17 & 18 & 19, 20 & 19, 20 \\
\hline
Test instances & 185 & 148 & 148 & 150 & 148 & 198 & 150 & 149 & 198 & 245 \\
\hline
Single Variable & 74.5\% & 73.5\% & 86.6\% & 84.1\% & 93.5\% & 89.0\% & 94.4\% & \textbf{85.2\%} & 47.7\% & 34.0\% \\
\hline
One Hot Vector & 60.9\% & 62.2\% & 84.0\% & 84.9\% & 86.4\% & 87.4\% & 88.0\% & 44.3\% & 35.0\% & 19.1\% \\
\hline
Population Code &\textbf{86.8\%} & \textbf{84.9\%} & \textbf{90.9\%} & \textbf{86.1\%} & \textbf{95.8\%} & \textbf{93.0\%} & \textbf{96.1\%} & 75.9\% & \textbf{82.6\%} & \textbf{74.0\%} \\

\hline\hline

\centering Object & \includegraphics[width=0.1\textwidth]{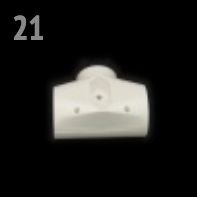} & \includegraphics[width=0.1\textwidth]{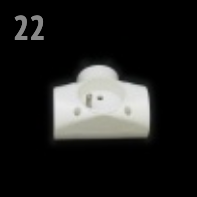} & \includegraphics[width=0.1\textwidth]{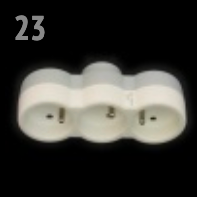} & \includegraphics[width=0.1\textwidth]{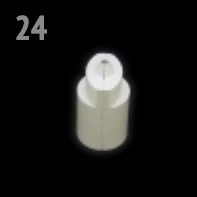} & \includegraphics[width=0.1\textwidth]{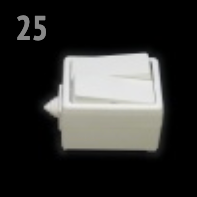} & \includegraphics[width=0.1\textwidth]{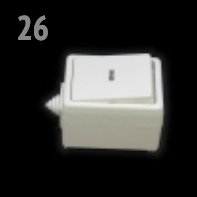} & \includegraphics[width=0.1\textwidth]{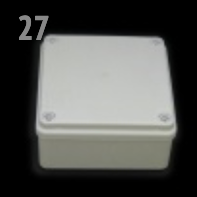} & \includegraphics[width=0.1\textwidth]{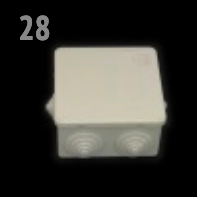} & \includegraphics[width=0.1\textwidth]{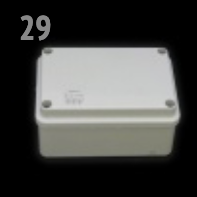} & \includegraphics[width=0.1\textwidth]{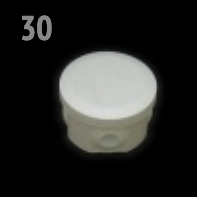} \\
\hline
Symmetries & None & None & 2 & Continuous & 2 & 2 & 4 & 2 & 2 & Continuous \\
\hline
Trained on Objects & 21, 22 & 21, 22 & 23 & 24 & 25, 26 & 25, 26 & 27 & 28 & 29 & 30 \\
\hline
Test instances & 184 & 195 & 250 & 195 & 98 & 100 & 96 & 195 & 99 & 148 \\
\hline
Single Variable & \textbf{62.7\%} & \textbf{64.0\%} & 66.4\% & 92.4\% & 75.9\% & 79.8\% & 7.8\% & 15.8\% & 18.6\% & 89.7\% \\
\hline
One Hot Vector & 34.5\% & 27.8\% & 34.6\% & 86.2\% & 41.3\% & 34.9\% & 17.8\% & 20.7\% & 31.2\% & 87.4\% \\
\hline
Population Code & 60.1\% & 59.7\% & \textbf{89.0\%} & \textbf{93.9\%} & \textbf{83.5\%} & \textbf{85.3\%} & \textbf{58.9\%} & \textbf{79.2\%} & \textbf{88.3\%} & \textbf{94.0\%}\\
\hline
\end{tabular}
}
\end{table*}

For training, we used the images from the BlenderProc4BOP dataset \cite{hodavn2020}, extracting all instances of an object with at least 60\% visibility. We augmented these data with the T-LESS Primesense camera images \cite{hodan2017}. For all images, we used the provided bounding boxes and made square cutouts by using the maximum of width and height plus a 10\% padding. The resulting cutouts were scaled to 128x128 pixels. We supplemented these images with 8,000 generated images per object using pyrender and the 3D model files from the T-LESS dataset. This combination resulted in about 20,000 to 22,000 images per object.

We combined some pairs of visually similar objects into one dataset and trained a single model for both objects (Tab. \ref{tab:all_obj}). This strategy resulted a small improvement (Tab. \ref{tab:abl}) and also points to the ability of our architecture and population-code representation to generalize across objects.

We trained all methods for 80 epochs, using the Adam optimizer with a learning rate of 0.0002 and a batch size of 64. As the loss function, we used MSE, except as specified differently above (Section IV.C). During training, we augmented the data for the pyrender-generated images by 1) adding random brightness, a uniformly-distributed value between -0.2 and 0.2 to all pixels with a 50\% probability and 2) adding a background image with a 70\% probability. Moreover, for all training images, we added Gaussian noise (std: 0.02) with a 50\% probability, did contrast normalization, and shifted the image in the image plane ($\pm 5$ pixels in x and y) with 90\% probability.

\begin{figure*}[t]
      \centering
      \includegraphics[width=0.7\textwidth]{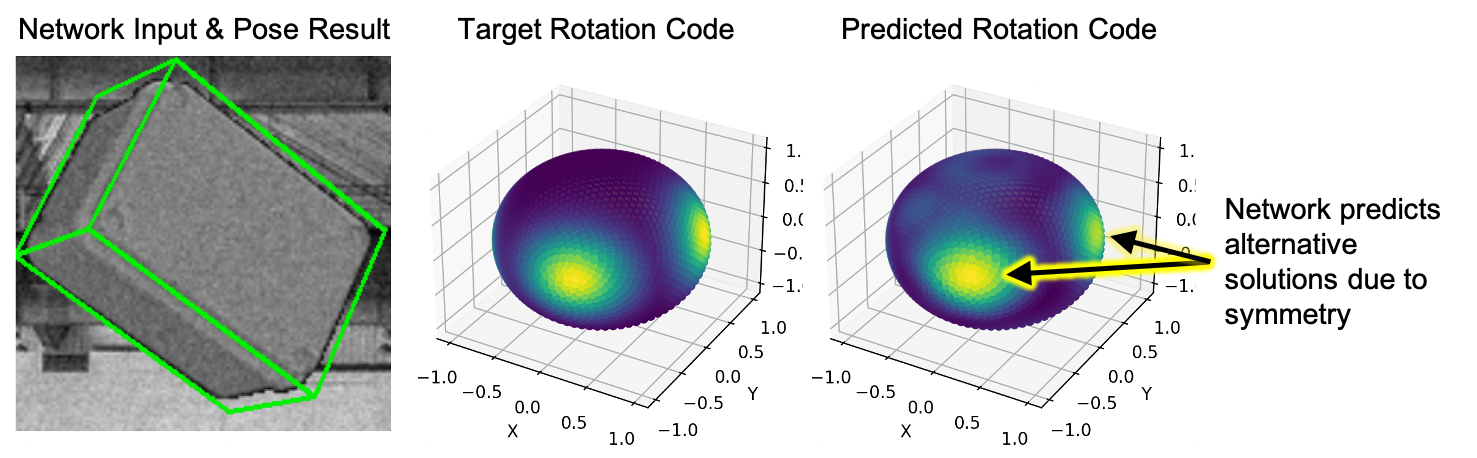}\vspace{-3mm}
      \caption{Our network learns the activations of the population code}\label{fig:learn_code}
 \end{figure*}

For testing, we used the official test set from the BOP Pose Estimation Challenge \cite{hodavn2020}. Since our focus was on rotation estimation rather than object detection, we used the provided ground truth bounding boxes. We compared Single Variable, One-Hot Vector, and Population Code across all 30 objects (Tab.~\ref{tab:all_obj}), whereas Multiple Pose Hypotheses was evaluated only on objects 4 and 5, as its performance was no better than random weights (Tab.~\ref{tab:obj4} and \ref{tab:obj5}). This suggests that the approach cannot learn from scratch and requires a pre-trained network backbone, as used by \cite{manhardt2019}.

For our method, the average inference time was {\bf 3.2 msec} on a Macbook Air M1 2020, which makes our method suitable for edge devices and real-time applications. In comparison, Multiple Pose Hypotheses required 115 msec for inference due to the costly clustering. 

For the population code, the  average MSSD accuracy across all objects was 84.7\% (Tab. \ref{tab:sym}). This value compares favorably against the best result on the BOP leaderboard for RGB-only input in the category {\it Localization of Seen Objects}: 81.4\%  (https://bop.felk.cvut.cz/leaderboards/pose-estimation-bop19/t-less/ accessed on 9/14/2024). The limitation is that we did not estimate the translation but instead used the ground truth translation.

Our average $e_{\mbox{\tiny VSD}}<0.3$ was 87.3\% across all objects (Tab. \ref{tab:sym}). In comparison, using the same ground-truth bounding boxes, Sundermeyer et al reported a 38.34\% accuracy from RGB input and 84.04\% with ICP refinement on the depth data \cite{sundermeyer2020}. However, the same caveat as above holds regarding estimating the translation.

Predicting a single variable performed well when there was no ambiguity in the target (Tab.~\ref{tab:sym}) but struggled with ambiguous poses (Tab.~\ref{tab:sym}). In contrast, our population code method performed equally well under ambiguity (Tab.~\ref{tab:sym}). Moreover, population code targets consistently outperformed one-hot vectors by large margins (Tab. \ref{tab:all_obj}).

We also observed that our network faithfully learned the entire distribution of the population code (Fig. \ref{fig:learn_code}) and not just the peaks. This behavior is expected since we compute the MSE across all neurons of a population code. Interestingly, even without providing symmetry information during training, the performance drop was relatively small (Tab. \ref{tab:abl}); i.e., the network could still learn to represent the ambiguity caused by symmetry.

\begin{table}[ht]
\caption{Accuracy depending on pose ambiguity caused by symmetry}\label{tab:sym}
\vspace{-1mm}
\centering
\begin{tabular}{|l||c|c|c|c|}
    \hline
    \multicolumn{5}{|c|}{\bf Without Ambiguity}\\ \hline
    Method          & VSD     & MSSD  & MSPD & $e_{\mbox{\tiny VSD}}<0.3$  \\ \hline \hline
    Single Variable                     & 73.41\% & 83.35\% & 85.74\% & 84.74\%\\ \hline
    One Hot Vector                      & 65.26\% & 73.27\% & 75.89\% & 78.81\%\\ \hline
    Population Code                      & \textbf{75.33\%} & \textbf{84.68\%} & \textbf{86.07\%} & \textbf{88.04\%}\\ \hline \hline
    \multicolumn{5}{|c|}{\bf With Ambiguity}\\ \hline
    Method          & VSD     & MSSD  & MSPD & $e_{\mbox{\tiny VSD}}<0.3$  \\ \hline \hline
    Single Variable                     & 44.70\% & 49.60\% & 43.51\% & 43.01\%\\ \hline
    One Hot Vector                      & 43.17\% & 40.12\% & 35.31\% & 43.04\%\\ \hline
    Population Code                      & \textbf{72.59\%} & \textbf{84.74\%} & \textbf{79.61\%} & \textbf{86.23\%}\\ \hline
    \multicolumn{5}{|c|}{\bf All Objects}\\ \hline
    Method          & VSD     & MSSD  & MSPD & $e_{\mbox{\tiny VSD}}<0.3$  \\ \hline \hline
    Single Variable                     & 61.77\% & 69.67\% & 68.62\% & 67.82\%\\ \hline
    One Hot Vector                      & 56.31\% & 59.83\% & 59.44\% & 64.31\%\\ \hline
    Population Code                      & \textbf{74.22\%} & \textbf{84.70\%} & \textbf{83.46\%} & \textbf{87.31\%}\\ \hline
\end{tabular}
\end{table}
  
\begin{table}[h]
\caption{Comparison of methods for object 4 in the T-LESS dataset}\label{tab:obj4}
\vspace{-1mm}
\centering
\resizebox{0.49\textwidth}{!}{
\begin{tabular}{|l||c|c|c|c|c|c|}
    \hline
    Approach              & VSD    & MSSD  & MSPD  &  $e_{\mbox{\tiny VSD}}<0.3$  &  VSS    & ADI \\ \hline \hline
    Random Weights  & 8.2\% & 9.0\% & 14.4\%  & 6.6\% & 55.5\% & 7.1\% \\ \hline 
    Multiple Poses      & 8.4\% & 6.3\% & 9.6\% & 2.6\% & 57.5\% & 4.9\%\\ \hline
    Single Variable     & 74.4\% & 84.1\% & 86.2\% & 85.0\% & 89.1\% & 87.1\%\\ \hline
    One-Hot Vector    & 68.5\% & 78.5\% & 80.6\%  & 81.6\% & 86.7\% & 82.6\%\\ \hline
    Population Code  & \bf{80.0\%} & \bf{86.8\%} & \bf{88.5\%} & \bf{89.5\%} & \bf{90.3\%} & \bf{90.5\%} \\ \hline
    \end{tabular}
}
\end{table}
 \begin{table}[h]
\caption{Comparison of methods for object 5 in the T-LESS dataset}\label{tab:obj5}
\vspace{-1mm}
\centering
\resizebox{0.49\textwidth}{!}{
\begin{tabular}{|l||c|c|c|c|c|c|}
    \hline
    Approach              & VSD    & MSSD  & MSPD  &  $e_{\mbox{\tiny VSD}}<0.3$ &   VSS   & ADI \\ \hline \hline
    Random Weights & 17.4\% & 1.7\% & 0.9\%     & 4.6\% & 68.0\% & 11.3\% \\ \hline 
    Multiple Poses     & 20.7\% & 3.3\% & 2.3\%.    & 9.8\% & 69.0\% & 30.4\% \\ \hline
    Single Variable     & 61.5\% & 66.0\% & 61.4\% & 73.2\% & 87.2\% & 82.5\% \\ \hline
    One-Hot Vector    & 69.3\% & 69.0\% & 66.6\%  & 82.5\% & 89.9\% & 87.1\% \\ \hline
    Population Code  & \bf{83.4\%} & \bf{89.6\%} & \bf{87.8\%} & \bf{94.9\%} & \bf{94.1\%} & \bf{96.9\%} \\ \hline
    \end{tabular}
}
\end{table}
  
\begin{table}[h]
\caption{Ablation results for our method using object 5 from T-LESS}\label{tab:abl}
\vspace{-1mm}
\centering
\resizebox{0.45\textwidth}{!}{
    \begin{tabular}{|p{3cm}||c|c|c|}
    \hline
    Ablation & VSD  & MSSD & MSPD  \\ \hline\hline
    Original\newline Trained on objects 5 \& 6                            & \bf{83.4\%} & \bf{89.6\%} & \bf{87.8\%} \\ \hline
    Trained only on object 5                                                         & 82.8\% & 88.2\% & 86.6\% \\ \hline
    Trained only on object 5,\newline removed 1x1 conv layer   &  81.4\% & 87.1\% & 84.7\% \\ \hline 
    Trained only on object 5,\newline trained without symmetry &  77.7\% & 84.9\% & 82.3\%\\ \hline 
    \end{tabular}
}
\end{table}

\section{CONCLUSIONS}

Representing object orientation with a population code allows end-to-end training of a pose-estimation network from scratch and solves the ambiguity problem caused by object symmetry. This strategy results in fast and accurate inference. The speed allows real-time operation on edge devices. We found that the population-code results in higher accuracies compared to other methods using the same training data and network. For robotic control, the population code could be also directly mapped onto a grasp posture (omitting rotation matrices), which we speculate is the operation used also by our brains.

\addtolength{\textheight}{-12cm}   









\bibliographystyle{IEEEtran}
\bibliography{literature}

\begin{thebibliography}{10}
\providecommand{\url}[1]{#1}
\csname url@samestyle\endcsname
\providecommand{\newblock}{\relax}
\providecommand{\bibinfo}[2]{#2}
\providecommand{\BIBentrySTDinterwordspacing}{\spaceskip=0pt\relax}
\providecommand{\BIBentryALTinterwordstretchfactor}{4}
\providecommand{\BIBentryALTinterwordspacing}{\spaceskip=\fontdimen2\font plus
\BIBentryALTinterwordstretchfactor\fontdimen3\font minus
  \fontdimen4\font\relax}
\providecommand{\BIBforeignlanguage}[2]{{%
\expandafter\ifx\csname l@#1\endcsname\relax
\typeout{** WARNING: IEEEtran.bst: No hyphenation pattern has been}%
\typeout{** loaded for the language `#1'. Using the pattern for}%
\typeout{** the default language instead.}%
\else
\language=\csname l@#1\endcsname
\fi
#2}}
\providecommand{\BIBdecl}{\relax}
\BIBdecl

\bibitem{Hoffmann2005}
H.~Hoffmann, \emph{Unsupervised Learning of Visuomotor Associations}, ser.
  {MPI} Series in Biological Cybernetics.\hskip 1em plus 0.5em minus
  0.4em\relax Logos Verlag Berlin, 2005, vol.~11.

\bibitem{wen2024}
B.~Wen, W.~Yang, J.~Kautz, and S.~Birchfield, ``Foundationpose: Unified 6d pose
  estimation and tracking of novel objects,'' in \emph{Proceedings of the
  IEEE/CVF Conference on Computer Vision and Pattern Recognition}, 2024, pp.
  17\,868--17\,879.

\bibitem{liu2024}
J.~Liu, W.~Sun, H.~Yang, Z.~Zeng, C.~Liu, J.~Zheng, X.~Liu, H.~Rahmani,
  N.~Sebe, and A.~Mian, ``Deep learning-based object pose estimation: A
  comprehensive survey,'' \emph{arXiv preprint arXiv:2405.07801}, 2024.

\bibitem{sundermeyer2020}
M.~Sundermeyer, Z.-C. Marton, M.~Durner, and R.~Triebel, ``Augmented
  autoencoders: Implicit 3d orientation learning for 6d object detection,''
  \emph{International Journal of Computer Vision}, vol. 128, no.~3, pp.
  714--729, 2020.

\bibitem{Hubel1959}
D.~H. Hubel and T.~N. Wiesel, ``\BIBforeignlanguage{eng}{Receptive fields of
  single neurones in the cat's striate cortex.}''
  \emph{\BIBforeignlanguage{eng}{J Physiol}}, vol. 148, no.~3, pp. 574--591,
  Oct 1959.

\bibitem{Pouget2000}
A.~Pouget, P.~Dayan, and R.~Zemel, ``Information processing with population
  codes,'' \emph{Nature Reviews Neuroscience}, vol.~1, no.~2, pp. 125--132,
  2000.

\bibitem{gilitschenski2019}
I.~Gilitschenski, R.~Sahoo, W.~Schwarting, A.~Amini, S.~Karaman, and D.~Rus,
  ``Deep orientation uncertainty learning based on a bingham loss,'' in
  \emph{International conference on learning representations}, 2019.

\bibitem{murphy2021}
K.~Murphy, C.~Esteves, V.~Jampani, S.~Ramalingam, and A.~Makadia,
  ``Implicit-pdf: Non-parametric representation of probability distributions on
  the rotation manifold,'' \emph{arXiv preprint arXiv:2106.05965}, 2021.

\bibitem{manhardt2019}
F.~Manhardt, D.~M. Arroyo, C.~Rupprecht, B.~Busam, T.~Birdal, N.~Navab, and
  F.~Tombari, ``Explaining the ambiguity of object detection and 6d pose from
  visual data,'' in \emph{Proceedings of the IEEE/CVF International Conference
  on Computer Vision}, 2019, pp. 6841--6850.

\bibitem{dixon91}
R.~A. Dixon, \emph{Mathographics}.\hskip 1em plus 0.5em minus 0.4em\relax Dover
  Publications, 1991.

\bibitem{liu2022}
Z.~Liu, H.~Mao, C.-Y. Wu, C.~Feichtenhofer, T.~Darrell, and S.~Xie, ``A convnet
  for the 2020s,'' in \emph{Proceedings of the IEEE/CVF conference on computer
  vision and pattern recognition}, 2022, pp. 11\,976--11\,986.

\bibitem{hodan2020}
T.~Hoda{\v{n}}, M.~Sundermeyer, B.~Drost, Y.~Labb{\'e}, E.~Brachmann,
  F.~Michel, C.~Rother, and J.~Matas, ``{BOP} challenge 2020 on {6D} object
  localization,'' in \emph{Computer Vision--ECCV 2020 Workshops: Glasgow, UK,
  August 23--28, 2020, Proceedings, Part II 16}.\hskip 1em plus 0.5em minus
  0.4em\relax Springer, 2020, pp. 577--594.

\bibitem{hodan2016}
T.~Hoda{\v{n}}, J.~Matas, and {\v{S}}.~Obdr{\v{z}}{\'a}lek, ``On evaluation of
  6d object pose estimation,'' in \emph{Computer Vision--ECCV 2016 Workshops:
  Amsterdam, The Netherlands, October 8-10 and 15-16, 2016, Proceedings, Part
  III 14}.\hskip 1em plus 0.5em minus 0.4em\relax Springer, 2016, pp. 606--619.

\bibitem{zhou2019}
Y.~Zhou, C.~Barnes, J.~Lu, J.~Yang, and H.~Li, ``On the continuity of rotation
  representations in neural networks,'' in \emph{Proceedings of the IEEE/CVF
  conference on computer vision and pattern recognition}, 2019, pp. 5745--5753.

\bibitem{wang2021}
G.~Wang, F.~Manhardt, F.~Tombari, and X.~Ji, ``{GDR-Net}: Geometry-guided
  direct regression network for monocular {6D} object pose estimation,'' in
  \emph{Proceedings of the IEEE/CVF Conference on Computer Vision and Pattern
  Recognition}, 2021, pp. 16\,611--16\,621.

\bibitem{hodan2017}
T.~Hodan, P.~Haluza, {\v{S}}.~Obdr{\v{z}}{\'a}lek, J.~Matas, M.~Lourakis, and
  X.~Zabulis, ``{T-LESS}: An {RGB-D} dataset for 6d pose estimation of
  texture-less objects,'' in \emph{2017 IEEE Winter Conference on Applications
  of Computer Vision (WACV)}.\hskip 1em plus 0.5em minus 0.4em\relax IEEE,
  2017, pp. 880--888.

\bibitem{hodavn2020}
T.~Hoda{\v{n}}, M.~Sundermeyer, B.~Drost, Y.~Labb{\'e}, E.~Brachmann,
  F.~Michel, C.~Rother, and J.~Matas, ``Bop challenge 2020 on 6d object
  localization,'' in \emph{Computer Vision--ECCV 2020 Workshops: Glasgow, UK,
  August 23--28, 2020, Proceedings, Part II 16}.\hskip 1em plus 0.5em minus
  0.4em\relax Springer, 2020, pp. 577--594.

\end{thebibliography}

\end{document}